\documentclass[letterpaper]{article} % DO NOT CHANGE THIS
\usepackage{aaai23}  % DO NOT CHANGE THIS
\usepackage{times}  % DO NOT CHANGE THIS
\usepackage{helvet}  % DO NOT CHANGE THIS 
\usepackage{courier}  % DO NOT CHANGE THIS
\usepackage[hyphens]{url}  % DO NOT CHANGE THIS  
\usepackage{graphicx} % DO NOT CHANGE THIS
\usepackage{natbib}  % DO NOT CHANGE THIS AND DO NOT ADD ANY OPTIONS TO IT
\usepackage{caption} % DO NOT CHANGE THIS AND DO NOT ADD ANY OPTIONS TO IT
\usepackage{algorithm}

\usepackage{epsfig}
\usepackage{graphicx}
\usepackage{amsmath}
\usepackage{mathtools}
\usepackage{amssymb}

\usepackage{siunitx}
\usepackage[figurename=Figure]{caption}
\usepackage{xspace}
\usepackage{multirow}
\usepackage{wrapfig}
\usepackage{amsthm}
\usepackage{thmtools}
\usepackage{subcaption}
\usepackage{algorithm}      % algorithm
\usepackage[noend]{algpseudocode} % algorithm
\usepackage[title]{appendix}
\usepackage{booktabs}
\usepackage[export]{adjustbox}
\usepackage{mathrsfs}

\renewcommand{\eqref}[1]{\mbox{Equation~(\ref{#1})}}

\def\Vec#1{{\boldsymbol{#1}}}

\def\Mat#1{{\boldsymbol{#1}}}

\newcolumntype{L}[1]{>{\raggedright\arraybackslash}p{#1}}
\newcolumntype{C}[1]{>{\centering\arraybackslash}p{#1}}
\newcolumntype{R}[1]{>{\raggedleft\arraybackslash}p{#1}}

\usepackage{newfloat}
\usepackage{listings}
\DeclareCaptionStyle{ruled}{labelfont=normalfont,labelsep=colon,strut=off} % DO NOT CHANGE THIS
\floatstyle{ruled}
\newfloat{listing}{tb}{lst}{}
\floatname{listing}{Listing}
\title{Text to Point Cloud Localization with Relation-Enhanced Transformer}
\author{
    Guangzhi Wang\textsuperscript{\rm 1}, 
    Hehe Fan\textsuperscript{\rm 2}, 
    Mohan Kankanhalli\textsuperscript{\rm 2}
}
\affiliations{
    \textsuperscript{\rm 1}Institute of Data Science, National University of Singapore \\
    \textsuperscript{\rm 2}School of Computing, National University of Singapore
    guangzhi.wang@u.nus.edu, 
    hehe.fan@nus.edu.sg,
    mohan@comp.nus.edu.sg
}

\usepackage{bibentry}
\begin{document}

\maketitle

\begin{abstract}
Automatically localizing a position based on a few natural language instructions is essential for future robots to communicate and collaborate with humans.
To approach this goal, we focus on the text-to-point-cloud cross-modal localization problem.
Given a textual query, it aims to identify the described location from city-scale point clouds.  
The task involves two challenges.    
1) In city-scale point clouds, similar ambient instances may exist in several locations.
Searching each location in a huge point cloud with only instances as guidance may lead to less discriminative signals and incorrect results. 
% Separately searching each location around the point of interest in a huge point cloud without considering their relations and then filtering mismatching areas increases the localization burden and may lead to incorrect results. 
2) In textual descriptions, the hints are provided separately. 
In this case, the relations among those hints are not explicitly described, leading to the difficulties of learning relations.
To overcome these two challenges, we propose a unified Relation-Enhanced Transformer (RET) to improve representation discriminability for both point cloud and natural language queries.
The core of the proposed RET is a novel Relation-enhanced Self-Attention (RSA) mechanism, which explicitly encodes instance (hint)-wise relations for the two modalities.    
Moreover, we propose a fine-grained cross-modal matching method to further refine the location predictions in a subsequent instance-hint matching stage.
Experimental results on the KITTI360Pose dataset demonstrate that our approach surpasses the previous state-of-the-art method by large margins. 

\end{abstract}
\section{Introduction}
\begin{figure}
    \centering
    \includegraphics[width=1.0\columnwidth, trim=0 0 0 0, clip]{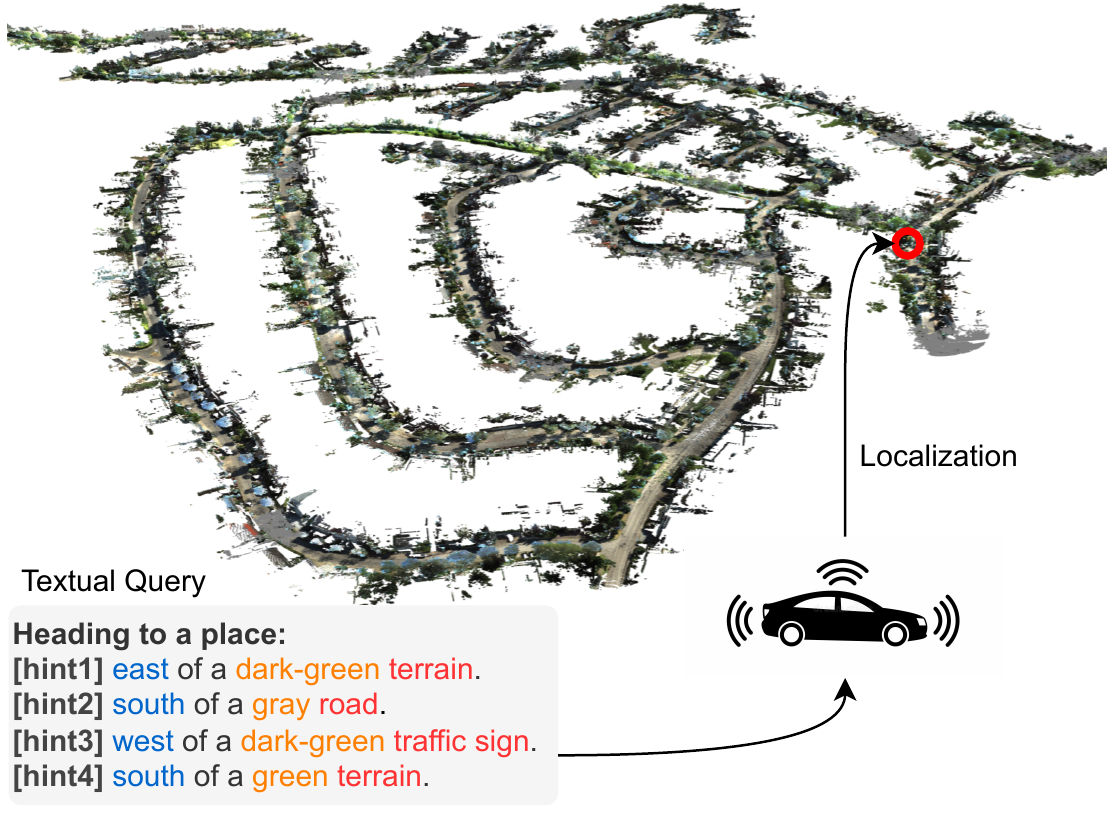}
    \vspace{-1em}
    \caption{Illustration of the text to point cloud localization task. Given a textual query, which usually contains several independent hints, the goal is to localize the point of interest in a huge city-scale point cloud.
    % The task of text to point cloud localization. Given a textual query, the agent is required to identify the described location from city-scale point cloud.
    }\label{fig:teaser}
\end{figure}

Understanding natural language instructions in the 3D real world is a fundamental skill for future artificial intelligence assistants to collaborate with humans.
In this paper, we focus on the outdoor environment and study the task of natural language-based localization from city-scale point clouds. 
As shown in Figure~\ref{fig:teaser},  given a linguistic description of a position, which contains several hints, the goal of the task is to find out the target location from a large-scale point cloud.
This task can effectively help mobile robots, such as self-driving cars and autonomous drones, cooperate with humans to coordinate actions and plan their trajectories. 
By understanding the destination from natural language instructions, it reduces the human effort required for manual operation.

% Understanding natural language instructions is an increasingly important skill for future artificial intelligence assistants to collaborate with humans.
% For example, autonomous robots that can understand human instructions will alleviate the operation efforts for humans.
% In this paper, we focus on outdoor robots, and study the task of natural language-based localization from large-scale point clouds.
% Given a linguistic description of a position, the agent is expected to identify the described location from city-scale point clouds.
% This task has huge application potential in modern outdoor robots, such as autonomous cars, unmanned aerial vehicles and delivery drones, which are expected to understand their destination from language instructions, thereby alleviating people's efforts for manual operation.

However, this task is intrinsically challenging.
Precise localization requires both correct language interpretation and effective large-scale point cloud understanding.
Considering the difficulties, an existing method~\cite{kolmet2022text2pos} first divides a city-wide point cloud into several cells, and then solves this task in a \textit{Coarse-to-Fine} manner.
% \HH{

The goal of the `coarse' stage is to find out the target cell that contains the queried location according to the given natural language descriptions.
In this stage, the instances included in point cloud cells and those mentioned in language descriptions are mainly used for text-to-point-cloud retrieval based on their types, without considering their relations. 
% instead of precise location prediction via their relations. 
% }
% for cell retrieval.  
% In the coarse stage, the scene point clouds are divided into cubic cells, each containing a small fraction of the scene.
% Based on the contained instances, the divided cells are encoded into vector representations, which are later retrieved by encoded textual descriptions.
% \HH{
In the `fine' stage, each object in the textual query is matched with an in-cell point cloud instance, whereby a target location will be predicted from each hint.
% that is associated with an instance.
% }
% according to the point cloud representations.
% The ultimate position is obtained by averaging the location predictions for all matched instances. 
% Afterwards, according to the description, the location is predicted for each matched instance. 
% and the final location is predicted based on matched instances.
% and an offset is predicted for each matched instance.
% The final location is then taken as the average of matched instance center plus the predicted position offset. 
This pioneering method sets up a significant starting point for tackling the challenging task. However, it fails to consider the intrinsic relations in both stages, resulting in sub-optimal performance.

% there are still unrecognized difficulties in both stages, prohibiting further advancement of this task.
For the coarse stage, because similar ambient instances may exist in several cells,  performing retrieval based on only the cell-contained and query-related instance types without considering their relations may lead to low discriminability for both cell and query representations, which inevitably leads to ambiguity.
Based on those low-discriminability representations, it is difficult to find out the correct cell. 
In the fine stage, we observe that insufficient \textit{cross-modal collaboration} leads to difficulties in location refinement.
Given the retrieved cell, precise location prediction requires joint understanding of both point clouds and textual queries.
% However, in the previous method~\cite{kolmet2022text2pos}, the cross-modal collaboration is insufficient or hampered in this process. 
However, in the previous method~\cite{kolmet2022text2pos}, the cross-modal collaboration is only performed from textual queries to point clouds in a single step, which results in optimization difficulty for multi-task learning.

In this work, we aim to solve the aforementioned shortcomings in both stages.
For the coarse stage, 
% even though instance-centric representations are less informative for both modalities, 
we propose to encode pairwise instance relations to improve representation discriminability for both modalities,
% need to ellaborate
which is achieved through a novel Relation-Enhanced Transformer (RET) architecture.
% particularly models instance relations. 
% explicitly encode pairwise instance relations. 
% through a Relation-enhanced Self-Attention (RSA) mechanism. 
In particular, the in-cell point cloud instance relations are modeled as their geometric displacements, while computed as the fusion of hint representations in the linguistic domain.
These relations from two modalities are respectively incorporated into their representation in a unified manner, 
which is achieved through the proposed Relation-enhanced Self-Attention (RSA) mechanism.
% resulting in more informative and discriminative representations for both modalities.
% which is finally encoded into visual and linguistic representations.
% Through the incorporation of instance relations, both the visual and linguistic representation are made more discriminative, resulting in more performant cell retrieval.  
For the fine stage, we perform Cascaded Matching and Refinement (CMR) to enhance cross-modal collaboration.
In particular, different from~\cite{kolmet2022text2pos} which achieves this objective in a single step, we perform description-instance matching and position refinement in two sequential steps.
Such formulation allows us to minimize the optimization difficulty of multi-objective learning and noisy intermediate results, thereby improving cross-modal collaboration. 

We validated the effectiveness of our method on the KITTI360Pose benchmark~\cite{kolmet2022text2pos}. 
Extensive experiments demonstrate that the proposed method can surpass the previous approach by a large margin, leading to new state-of-the-art results. Our contributions are three-fold:
% We also perform additional ablation studies to prove the effectiveness of each component in our method.
% Additional qualitative results are also provided for further investigation of the proposed method.
% which computes value from 
% To summarize, our contributions are three-fold:
\begin{itemize}
    \item We propose a novel Relation-Enhanced Transformer (RET) to improve representation discriminability for both point clouds and textual queries.
    The core component of RET is the Relation-enhanced Self-Attention (RSA) mechanism, which
    encodes instance (hint) relations for the two modalities in a unified manner.
    % so as to improve the cross-modal retrieval performance.
    \item We propose to perform cross-modal instance matching and position refinement in two sequential steps. 
    This formulation allows us to minimize the optimization difficulty of multi-task learning and the influence of noisy intermediate results, thereby improving cross-modal collaboration for fine-grained location prediction.
    \item We perform extensive experiments on the KITTI360Pose dataset ~\cite{kolmet2022text2pos}.
    The results show that our approach can surpass previous method by a large margin, resulting in new state-of-the-art performance.
    Additional ablation studies further demonstrate the effectiveness of each  component in the proposed method.
    % \footnote{Our source code is provided in the supplementary material and will be released to facilitate future research.}.
\end{itemize}
\section{Related Work}
\noindent\textbf{Transformer and Attention Mechanism.}
Transformer and self-attention mechanism~\cite{vaswani2017transformer,fan21p4transformer} has become increasingly popular in recent years.
Although first proposed for natural language processing, with architectural adaptation, Transformer has been widely applied to many vision tasks including visual recognition~\cite{dosovitskiy2020image, liu2021swin}, object detection~\cite{carion2020end, zhu2020deformable} and semantic segmentation~\cite{cheng2021per}.
Besides, the transformer-based architectures are also utilized to model cross-modal (\textit{e.g.}, vision and language) relations~\cite{tan2019lxmert, lu2019vilbert,Li_2019_ICCV, zhang_moment_sigir, li2022blip}.
In these architectures, the attention mechanism is widely employed to implicitly learn relations among the input tokens.
Nevertheless, without explicit relation encoding, the vanilla Transformer can only encode relations implicitly with the help of positional encoding~\cite{dosovitskiy2020image}. % or similarity learning.
To facilitate better relation modeling, some works modulate the attention computation process by explicitly incorporating element relations. 
For example, ~\cite{wu2021rethinking} modified the attention mechanism via unified relative position bias to improve visual recognition.
For object detection, spatial relations between bounding boxes are introduced to modulate the attention weights~\cite{liu2022dabdetr, Gao_2021_ICCV_SMCA}.
For dynamic point cloud analysis, displacement between points \cite{fan2022point} is utilized for point-specific attention computation. 
In this work, we propose to model relations for both point clouds and language queries by explicitly incorporating intra-modality relations in a unified manner.

\noindent\textbf{Visual Localization.}
The task that is most related to ours is vision-based localization~\cite{arandjelovic2016netvlad, brachmann2017dsac, hausler2021patch}, which is to estimate a pose based on an image or image sequence.
Existing methods mostly solve this task in two stages~\cite{sarlin2019coarse, sattler2016efficient, zhou2020learn}. 
The first stage finds a subset of all images using image retrieval-based techniques~\cite{arandjelovic2016netvlad, hausler2021patch, torii201524}, while the second stage establishes pixel-wise correspondence between the query image and the retrieved one to predict the precise pose.
In this work, we also study the task of localization in a coarse-to-fine manner, but differ from visual localization in that: 1) we try to infer the location from city-wide point clouds instead of images. 2) we try to estimate the pose from textual query rather than images.
Compared to visual localization, our task requires multi-modal understanding and is more challenging to solve.

\begin{figure*}
    \centering
    \includegraphics[width=1.0\textwidth]{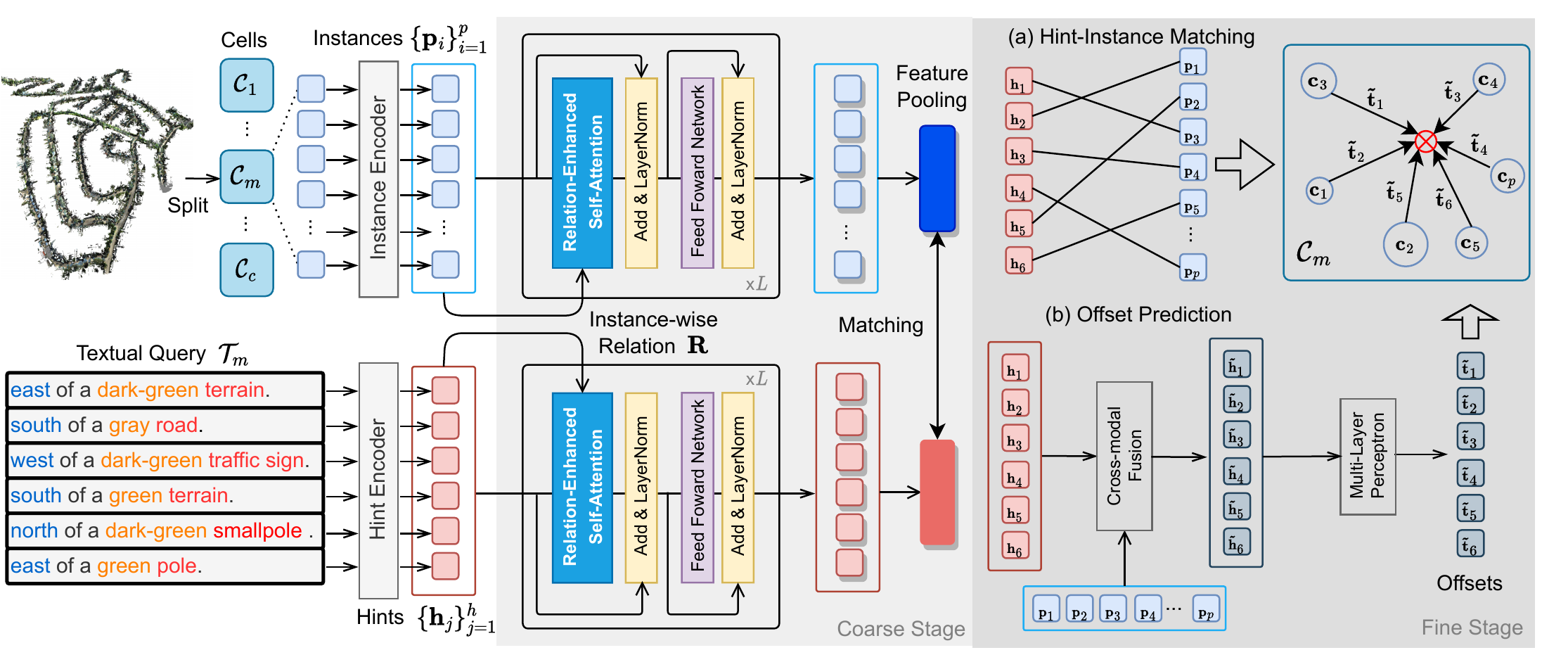}
    \caption{Framework of the proposed method. The city-scale point cloud is first divided into individual cells. 
    Then, in the coarse stage, the cells and the textual query are respectively encoded with the proposed Relation-Enhanced Transformer (RET), which are later used for query-cell matching.
    In the fine stage, each hint is matched with an in-cell instance. 
    Then, cross-modal fusion dynamically aggregates hints and instance representations for offset prediction. 
    The target location is predicted based on matching results and offset predictions.}
    \label{fig:framework}
\end{figure*}

\noindent\textbf{3D Language Grounding.}
As we humans live in a 3D world and communicate through natural language, recent work has begun to investigate the tasks on the cross-modal understanding of 3D vision and natural language.
Among these tasks, the one that is most related to ours is 3D language grounding, which aims at localizing an object in point clouds from a given natural language query.
For example, ScanRefer~\cite{chen2020scanrefer} studies 3D language grounding from real-life in-door scenes. 
ReferIt3D~\cite{achlioptas2020referit3d} studies a related task under a simpler setting, which assumes the object instances are segmented in advance.
% except that it assumes the object instance segmentation is given in advance. 
InstanceRefer~\cite{yuan2021instancerefer} improves previous methods by adopting a 3D panoptic segmentation backbone, utilizing multi-level visual context.
Recently, graph structure~\cite{feng2021free} is also utilized to improve the representation learning qualities.

\section{Methodology}

\subsection{Preliminaries}
Given a textual query, our goal is to identify the position it describes from a city-scale point cloud.
To handle the large-scale point cloud, we divide each scene into a set of cubic cells of fixed size by a preset stride.
% \footnote{}.
Each cell $\mathcal{C}$ contains a set of $p$ point cloud instances, which are encoded by PointNet++~\cite{qi2017pointnet++} into vector representations $\{\Vec{p}_i\}_{i=1}^{p}$.
Following~\cite{kolmet2022text2pos}, the textual query $\mathcal{T}$ is represented as a set of hints $\{\Vec{h}_j\}_{j=1}^{h}$, each encoding the direction relation between the target location and an instance.

Inspired by the existing work~\cite{kolmet2022text2pos}, given the cell splits, we solve this task in a coarse-to-fine manner with two stages.
The coarse stage is formulated as textual query based cell retrieval.
The goal of this stage is to train a model that encodes $\mathcal{C}$ and $\mathcal{T}$ into a joint embedding space whereby matched query-cell pairs are close while those unmatched are pulled apart~\cite{kiros2014unifying}.
In the fine stage, given a retrieved cell, we aim to refine the position prediction by utilizing fine-grained cross-modal information.
In particular, we first match each hint in the query with an in-cell instance by formulating it as an optimal transport problem~\cite{Liu_2020_CVPR}. 
After that, with the matching results, we predict the target location through a cross-modal fusion of point cloud instance and hint representations.
Based on the fused representation, we predict the target location for each matched instance.
% as a regression task.
Finally, we obtain the target location prediction based on a weighted combination of the matching and location prediction results.
The framework of our method is shown in Figure~\ref{fig:framework}.
In the following of this section, we will explain the proposed method for coarse stage and fine stage. 
% in Section~\ref{method:coarse} and Section~\ref{method:fine} respectively.
After that, our training and inference procedure will be detailed.

\subsection{Coarse Stage: Relation-Enhanced Transformer}\label{method:coarse}
After the cell split, the goal of the coarse stage is to successfully retrieve the cell $\mathcal{C}$ given a textual query $\mathcal{T}$.
To approach this objective, we need to encode $\mathcal{C}$ and $\mathcal{T}$ into a joint embedding space.
An intuitive solution is to encode both $\mathcal{C}$ and $\mathcal{T}$ based on the instances they contained as is done in~\cite{kolmet2022text2pos}.
However, with such representations, the low discriminability for cells and textual queries results in poor retrieval performance.
We argue that this can be attributed to the following two reasons.
On the one hand, the outdoor scenes are often of low diversity, whereby a group of mentioned instances can appear at multiple different locations.
% Different locations often have similar surrounding instances. 
Thus, simply describing a cell with its contained instances can result in less discriminative representations.
On the other hand, the textual queries often contain limited clues compared to the point clouds, making this cross-modality retrieval especially challenging. 
To this end, we propose to explicitly encode instance-relations to provide more discriminative representations for both modalities.
% The motivation behind our idea is that the relations between a certain instance and other instances make it different from other instances within the same class.

\begin{figure}
    \centering
    \includegraphics[trim=0 10 0 0, clip]{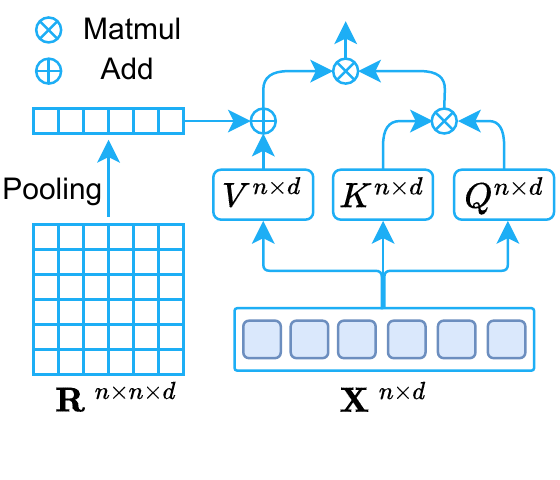}
    \vspace{-1em}
    \caption{Illustration of the proposed Relation-enhanced Self-Attention (RSA) mechanism. Pairwise relations are explicitly encoded into the value computation process.}\label{fig:rsa}
\end{figure}

% Transformer is used to model relations
The Transformer~\cite{vaswani2017transformer} has been widely utilized for relation-based representation learning in various tasks~\cite{Hu_2018_CVPR_relationnet, liu2021swin, fan2022point}.
The key component of the Transformer is the Self-Attention (SA) operation:
\begin{equation}
    \texttt{Attn}(\Mat{Q}, \Mat{K}, \Mat{V}) = \texttt{Softmax}(\Mat{Q}\Mat{K}^{T}/\sqrt{d})\Mat{V},
\end{equation}
where $d$ is the representation dimension and $\Mat{Q}, \Mat{K}, \Mat{V} \in \mathbb{R}^{N \times d} $ are the query, key and value matrices by transforming in-cell instances (or hints for textual queries) with corresponding linear transformations:
\begin{equation}
    \Mat{Q} = \Mat{W}^Q\Mat{X}, \Mat{K}=\Mat{W}^K\Mat{X}, \Mat{V}=\Mat{W}^V\Mat{X},
\end{equation}
with $\Mat{W}^* \in \mathbb{R}^{d\times d}$ are learnable matrices and $\Mat{X} = \Mat{P} \in \mathbb{R}^{p \times d}$ or $\Mat{H} \in \mathbb{R}^{h \times d}$ represents stacked instances\footnote{Note that the attention operation is often performed in different subspaces with multiple heads, which is omitted for simplicity.}.

Despite its generality, the vanilla SA lacks explicit relations in both modalities, thus is less informative to represent the cell and query. 
% that is easy to obtain but difficult to learn.
% relations modeling and thus less informative.
To this end, we propose a novel Relation-Enhanced Transformer (RET) to model explicit instance relations in both point clouds and textual descriptions.
Our RET is a stack of multiple Transformer encoder layers, except that, in place of SA, we propose a Relation-enhanced Self-Attention (RSA) to explicitly incorporate relation information into value computation. 
The computation process is shown as follows and illustrated in Figure~\ref{fig:rsa}.
\begin{equation}
% \begin{split}
\small
    \texttt{RSA}(\Mat{Q}, \Mat{K}, \Mat{V}, \Mat{R}) = \\ \texttt{Softmax}(\Mat{Q}\Mat{K}^T/\sqrt{d})(\Mat{V} + \texttt{Pool}(\Mat{R}, 1)),
% \end{split}
\end{equation}
where $\Mat{R} \in \mathbb{R}^{N \times N \times d}$ captures pairwise relations with $\Mat{R}_{ij} \in \mathbb{R}^{d}$ representing the relation between the $i$-th and $j$-th instance (hint).
$\texttt{Pool}(\Mat{R}, 1)$ indicates pooling tensor $\Mat{R}$ along dimension $1$.
In this way, our model can explicitly encode instance relations through this computation process, leading to more informative representations. 

The definition of relation varies flexibly with task objective and input modality.
% For text-based point cloud retrieval, direction relations between two instances play an important role for cell retrieval as 
For point cloud data, we take the geometric displacement of two instances as their relations, as direction is often mentioned in textual queries and thus informative for retrieval:\footnote{We have also tried other features such as number of points and bounding boxes of instances but didn't observe performance improvement.}
\begin{equation}
    \Mat{R}_{ij}^V = \Mat{W}^V(\Vec{c}_i - \Vec{c}_j),
\end{equation}
where $\Vec{c}_i \in \mathbb{R}^3$ represents the center coordinate of the $i$-th instance and $\Mat{W}^v \in \mathbb{R}^{d \times 3}$ transforms the displacement into embedding space.
For the linguistic description, 
% the direction relations can not be obtained directly as spatial features.
we compute the hint relation as the concatenation of their embeddings:
\begin{equation}
    \Mat{R}^{L}_{ij} = \Mat{W}^L[\Vec{h}_i ; \Vec{h}_j], 
\end{equation}
where $\Mat{W}^L \in \mathbb{R}^{d \times 2d}$ transforms the linguistic feature into representation space.
With the computation of RSA, the instance-wise relations for different modalities can be uniformly incorporated into query or cell representations
% .in a unified manner.

Finally, the cell (description) representations $\mathcal{C}_m$ ($\mathcal{T}_m$) are obtained via a pooling operation over all instances (hints) output from the RET for cross-modal retrieval.

\begin{table*}[]
    \centering
    \small
    \caption{Performance comparison on the KITTI360Pose.}\label{tab:valid_results}
    \vspace{-1em}
    \begin{tabular}{l|ccc|ccc}
    \toprule
    \multirow{3}{*}{Method} & \multicolumn{6}{c}{Localization Recall ($\epsilon < 5/10/15m $) $\uparrow$} \\ \cline{2-7}
    & \multicolumn{3}{c|}{Validation Set} & \multicolumn{3}{c}{Test Set} \\  \cline{2-7}
      & $k=1$ & $k=5$ & $k=10$ & $k=1$ & $k=5$ & $k=10$\\
    %  \midrule
    % Random & 0.00/0.01/0.01 & 0.01/0.02/0.03 & 0.02/0.04/0.06 & 0.00/0.00/0.01 & 0.01/0.01/0.02 & 0.01/0.02/0.03 \\
    \midrule
    Text2Pos~\cite{kolmet2022text2pos}     & 0.14/0.25/0.31 & 0.36/0.55/0.61 & 0.48/0.68/0.74  & 0.13/0.21/0.25 & 0.33/0.48/0.52 & 0.43/0.61/0.65 \\
    RET (Ours)     & \textbf{0.19}/\textbf{0.30}/\textbf{0.37} & \textbf{0.44}/\textbf{0.62}/\textbf{0.67} &\textbf{0.52}/\textbf{0.72}/\textbf{0.78}   & \textbf{0.16}/\textbf{0.25}/\textbf{0.29} & \textbf{0.35}/\textbf{0.51}/\textbf{0.56} &  \textbf{0.46}/\textbf{0.65}/\textbf{0.71} \\
    % \midrule 
    % $+\Delta$ & 0.05/ & & \\
    \bottomrule
    \end{tabular}

\end{table*}

\subsection{Fine Stage: Cascaded Matching and Refinement}\label{method:fine}
Following the coarse stage, we aim to refine the location prediction within the retrieved cell in the fine stage.
% ~\cite{nan2020reasoning}.
% the objective of the fine stage is to infer the precise location within the cell from the language descriptions. 
% Previous method~\cite{kolmet2022text2pos} matches each hint $\Vec{h}_j$ with an instance $\Vec{p}_i$ in the cell.
Inspired by~\cite{kolmet2022text2pos}, we perform instance matching and location refinement to utilize the fine-grained visual and linguistic information, which involves the following two objectives:
(1) For each hint, we find the in-cell instance it refers to via a matching process.
(2) For each matched pair $(i, j)$, a regressor predicts an offset $\Vec{\hat{t}}_i \in \mathbb{R}^2$ for each matched hint $\Vec{h}_j$, which represents the offset from the instance center $\Vec{c}_i$ to the target location.\footnote{For position prediction, we ignore the height information and considers 2D coordinates only.}
% The final position $\Vec{g}$ is predicted as the average of predictions from matched instances:
% \begin{equation}
% \Vec{\hat{g}} = \texttt{Average}(\Vec{c}_i + \Vec{t}_i),                   
% \end{equation}
% where $\Vec{\hat{t}}_i$ = $\texttt{MLP}(\Vec{h}_j)$ for each matched $(i, j)$.
% \subsubsection{Hint-Instance Matching}
% Optimal Matching...

% We found such a solution mainly suffers from two drawbacks. 
Previous method~\cite{kolmet2022text2pos} achieves the two objectives within a single step. 
However, given the objective of both hint-instance matching and offset prediction, the multi-task learning process introduces optimization difficulty.
Furthermore, in the early training steps,
the matcher is only partially trained, which produces noisy matching results.
The regressor learns and makes predictions based on this noisy results, leading to unstable learning process and sub-optimal performance.
% the noisy matching result in the early learning stage introduces difficulty of the offset prediction task, leading to sub-optimal position prediction.
% Especially, in the early learning stage, the sub-optimal matching results given by the matcher that is yet to be optimized introduces noise for offset regression.  
% , as well as noisy objective for the latter task.
% Secondly, the offset prediction is based only on the instance, with geometry information fully ignored.

To this end, we propose a Cascaded Matching and Refinement (CMR) strategy for the fine stage, where hint-instance matching and offset regression are sequentially performed. 
Specifically, following~\cite{kolmet2022text2pos}, we first train the SuperGlue~\cite{sarlin2020superglue} matcher for hint-instance matching, which is formulated as an optimal-transport problem.
Given the trained matcher, we obtain a set of hint-instance matching results $\{\Vec{p}_i, \Vec{h}_j, w_i\}_{j=1}^{h}$, where $w_i$ represents the confidence of the match. 
Then, to reduce the noise for regression, we predict the target location according to matched instances only.
% based on cross-modality information.

Precise location prediction requires proper understanding on both point cloud (what and where the referred instance is, \textit{e.g.}, \texttt{dark-green terrain}) and language description (what is the relation between the matched instance and the target location, \textit{e.g.}, \texttt{east of}). 
% Furthermore, comprehensive understanding of the cell instances is also necessary 
% (positions with many instances are less likely to be the target).
For this, we propose to facilitate cross-modal collaboration via the Cross-Attention (CA) mechanism, which is commonly used for cross-modality information fusion.
\begin{equation}
    \texttt{\texttt{CA}}(\Mat{H}, \Mat{P}) = \texttt{Attn}(\Mat{W}^{Q}\Mat{H}, \Mat{W}^{K}\Mat{P}, \Mat{W}^{V}\Mat{P}),
\end{equation}
where $\Mat{H}$, $\Mat{P}$ represent hints and instances, respectively, and $\Mat{W}^{*}$ are learnable transformation matrices.
Shortcut connection and layer normalization~\cite{ba2016layernorm} follows the cross-attention operation.
With these operations, the hint representation $\Vec{h}_i$ is accordingly updated to $\Vec{\tilde{h}}_i$ by dynamically fusing visual information.
As such, the information in the two modalities are joint utilized with the help of cross-modal collaboration.

Then, we predict the offset (the direction vector from instance center to target location) from the updated hint:
\begin{equation}
\Vec{\hat{t}}_i = \texttt{MLP}(\Vec{\tilde{h}}_j).
\end{equation}
To utilize the matching results, the final prediction is obtained via a weighted combination of each hint's prediction:
\begin{equation}\label{eq:prediction}
    {\Vec{\hat{g}}} = \sum_{i} \frac{w_i}{\sum_{m}w_m} (\Vec{c}_i + \Vec{\hat{t}}_i),
\end{equation}
where $w_i \in [0, 1]$ is the confidence score of the match $(\Vec{p}_i, \Vec{h}_j, w_i)$ and is set to $0$ for non-matched instances.
To filter out noisy matches, we consider only matches with confidence score greater than 0.2. 

% \subsubsection{Geometry-Aware Position Refinement}

\subsection{Training and Inference}\label{method:train}
\noindent\textbf{Training.}
For the coarse stage, we train the proposed RET for cross-modal retrieval with pairwise ranking loss~\cite{kiros2014unifying}:
\begin{equation}
\begin{split}
    \mathcal{L}_{coarse} &= \sum_{m=1}^{N_b}\sum_{n\neq m}^{N_b}
    [\alpha - \langle\mathcal{C}_m,\mathcal{T}_m\rangle + \langle\mathcal{C}_m, \mathcal{T}_n\rangle]_+ \\
    &+ \sum_{m=1}^{N_b}\sum_{n\neq m}^{N_b}
    [\alpha - \langle\mathcal{T}_m,\mathcal{C}_m\rangle + \langle\mathcal{T}_m, \mathcal{C}_n\rangle]_+, 
\end{split}
\end{equation}
where $N_b$ is the batch size, $\alpha$ is a hyper-parameter to control the separation strength and $\langle\cdot,\cdot\rangle$ represents inner product between vectors.
This loss function encourages the representation of matched description-cell pair to be by a margin $\alpha$ closer than those unmatched.
For the fine stage, we employ the loss in~\cite{sarlin2020superglue} to train the matcher, while $L_2$ loss is applied to train the offset regressor.

\noindent\textbf{Inference.}
% Given a textual query, we encode it with the proposed Relation Transformer and retrieve top-$k$ cells with the highest similarity. 
We first encode all cells and queries into a joint embedding space with the proposed Relation-Enhanced Transformer.
Then, for each query representation, we retrieve top-$k$ cells with highest similarity.
For each retrieved cell, we use the SuperGlue matcher trained in the fine stage to match each hint with an in-cell instance, which is followed by offset prediction based on the fused representations.
Finally, the position prediction is given by Eq.~\ref{eq:prediction}.
\section{Experiments}

\subsection{Dataset and Implementation Details}
% \vspace{-0.5em}
\noindent\textbf{Dataset Details.}
We evaluate our method on the recently proposed \textit{KITTI360Pose} dataset~\cite{kolmet2022text2pos}, which is built upon the KITTI360 dataset~\cite{Liao2021ARXIV_kitti360} with sampled locations and generated hints.
It contains point clouds of a total of 9 scenes, covering 14,934 positions with a total area of 15.51$km^2$.
We follow~\cite{kolmet2022text2pos} to use five scenes for training, one for validation, and the remaining three for testing.
% \noindent\textbf{Dataset Preprocessing.}
We sample the cells of size 30m with a stride of 10m.
% For each scene, the point clouds are splitted into cells of size $30m$ with a stride of $10m$.
For more details on the dataset preprocessing, please refer to our supplementary material.

\noindent\textbf{Implementation Details}
For the coarse stage, we trained the model with AdamW optimizer~\cite{loshchilov2018decoupled_adamw} with a learning rate of 2e-4. 
The models are trained for a total of 18 epochs while the learning rate is decayed by 10 at the 9-th epoch.
The $\alpha$ is set to 0.35.
For the fine stage, we first train the matcher with a learning rate of 5e-4 for a total of 16 epochs.
Afterwards, we fix the matcher and train the regressor based on the matching results for 10 epochs with a learning rate of 1e-4.
The regressor is formulated as a 3 layer Multi-Layer Perceptron.
Both of the two steps adopt an Adam~\cite{kingma2014adam} optimizer. 
The RET has 2 encoder layers for both point cloud part and linguistic part, each utilizing the Relation-enhanced Attention (RSA) mechanism with 4 heads and hidden dimension 2048.
For the two stages, we encode each instance in the cell with PointNet++~\cite{qi2017pointnet++} provided by Text2Pos~\cite{kolmet2022text2pos} for a fair comparison.
The hint representations are obtained by concatenating learned word embeddings.
More details are provided in our appendix.\footnote{Code available at: \url{https://github.com/daoyuan98/text2pos-ret}}
% \vspace{-0.5em}

\subsection{Comparison with the State-of-the-art}
% \vspace{-0.5em}
We compared our method with Text2Pos~\cite{kolmet2022text2pos} on the KITTI360Pose dataset. 
Following~\cite{kolmet2022text2pos}, we report top-$k$ ($k = 1/5/10$) recall rate of different error ranges $\epsilon < 5/10/15m$ for comprehensive comparison.
The results are shown in Table~\ref{tab:valid_results}.
% It can be seen that, when we randomly select the cells, the performance is close to zero with $k=1$, showing the difficulty of this task.
Text2Pos gives a recall of 0.14 when $k=1$ and $\epsilon < 5m$.
In contrast, our method can significantly improve the recall rate to 0.19, which amounts to $35.7\%$ relative improvement upon the baseline. 
Furthermore, when we relax the localization error constraints or increase $k$, consistent improvements upon the baseline can also be observed.  
For example, with $\epsilon < 5m$, our method achieves top-5 recall rate of $0.44$, which is $8\%$ higher than previous state-of-the-art.
Similar improvements can also be seen on the test set, showing our method is superior to the baseline method.

\subsection{Ablation Studies}
In this section, we perform ablation studies for both stages to investigate the effectiveness of each proposed component in our method.
The ablation studies for coarse stage and fine stage are provided separately for clear investigation.

% \vspace{0.5em}
\noindent\textbf{Coarse Stage.}
We study the importance of explicit relation incorporation in the coarse stage.
Since the coarse stage is formulated as a retrieval task, we use top-1/3/5 recall rate as evaluation metric, whereby the cell that contains the ground truth location is defined as positive.
% at of error range $\epsilon<5m$ to evaluate the retrieval performance for detailed comparison.
% A cell is defined as positive if it con

\noindent\textbf{\textit{Relation Incorporation.}}
We first study the necessity of explicit relation modeling for both point cloud and textual queries. 
The results are shown in Table~\ref{tab:exp_ret}.
It can be observed that relation modeling contributes significantly to successful retrieval.
In particular, without any relation incorporation, the top-5 recall rate is 0.32.
With the explicit fusion of linguistic relation, we observe an increase of 0.05 recall rate under same condition.
Besides, with the incorporation of visual (point cloud instance) relations only, the top-5 recall rate can be improved by 0.08, indicating explicit relations in the point clouds play a more important role.
Finally, with both relations, we achieve an improvement of 0.12 at top-5 recall rate upon that without any relation, showing that both visual and linguistic relations are necessary and complementary to improve the cell retrieval performance.
% We also followed~\cite{kolmet2022text2pos} to provide the results at different error threshold ($\epsilon < 5/10/10m$).

\begin{table}[]
    \centering
    \setlength{\tabcolsep}{9.0pt}
    \small
    \caption{Ablation study of the Relation-Enhanced Transformer (RET) on KITTI360Pose validation set. "wo X relation" indicates replacing the proposed RSA with the vanilla Self-Attention in corresponding modality.}\label{tab:exp_ret}
    \vspace{-1em}
    \begin{tabular}{l|ccc}
    \toprule
    %  & \multicolumn{3}{c}{Localization Recall ($\epsilon < 5m$) $\uparrow$} \\
    Method & $k=1 \uparrow$ & $k=3 \uparrow$ & $k=5 \uparrow$ \\
    \midrule
    w/o both relations      & 0.11 & 0.24 & 0.32 \\
    w/o linguistic relation     & 0.14 & 0.28 & 0.37  \\
    w/o visual relation & 0.16 & 0.30 & 0.40 \\
    \midrule
    Full (Ours) & \textbf{0.18} & \textbf{0.34} & \textbf{0.44} \\
    \bottomrule
    \end{tabular}
\end{table}

\noindent\textbf{\textit{RET Hyper-parameters.}}
We also studied the importance of the hyper-parameters involved in RET, namely the number of layers of RET and the number of heads of RSA.
The results are shown in Table~\ref{tab:exp_layers}.
It can be observed that, thanks to the strong relation modeling capacity of the proposed RET, we can obtain the best performance with 2 layers and 4 heads in the RSA.
Decreasing and increasing the number of layers both lead to worse performance, which may be attributed to underfitting and overfitting, respectively.

\begin{table}[]
    \centering
    \small
    \caption{The effects of \#layers of RET and \#heads of RSA.}\label{tab:exp_layers}
    \vspace{-1em}
    \setlength{\tabcolsep}{9.0pt}
    \begin{tabular}{ccccc}
    \toprule
    % & & \multicolumn{3}{c}{Localization Recall ($\epsilon < 5m$) $\uparrow$}\\
    \#Layers & \#Heads  &  $k=1 \uparrow$ & $k=3 \uparrow$ & $k=5 \uparrow$ \\
    \midrule
    % 1     &  2 & & & \\
    1     &  4  & 0.16 & 0.31 & 0.40 \\
    1     &  8  & 0.16 & 0.30 & 0.40 \\
    \midrule
    2     &  2  & 0.17 & 0.32 & 0.42 \\
    2     &  4  & \textbf{0.18} & \textbf{0.34} & \textbf{0.44} \\
    2     &  8  & 0.16 & 0.31 & 0.40 \\
    \midrule
    3     &  4  & 0.16 & 0.32 & 0.39 \\
    3     &  8  & 0.15 & 0.29 & 0.37 \\
    \bottomrule
    \end{tabular}

\end{table}
% \vspace{-2em}

% \vspace{0.5em}
\noindent\textbf{Fine Stage.}
The objective of the fine stage is to correctly \textit{match} linguistic hints and point cloud instances and \textit{regress} the target location.
Thus, we study the performance of the \textit{matcher} and \textit{regressor}, respectively. 
% , given the ground truth cell, we adopt the regression error to evaluate the regression capacity of the model in the fine stage.

\begin{table}[]
    \centering
    \small
    \caption{Comparison of training strategy and matcher performance on the KITTI360Pose dataset.}\label{tab:exp_matcher}
    \vspace{-1em}
    \begin{tabular}{l|cc|cc}
    \toprule
     \multirow{2}{*}{Strategy} & \multicolumn{2}{c|}{Train} & \multicolumn{2}{c}{Validation} \\  \cline{2-5}
     & Precision $\uparrow$ & Recall $\uparrow$ & Precision $\uparrow$ & Recall $\uparrow$ \\
    \midrule
    joint     &  98.12 & 98.16 & 86.67 & 87.59 \\
    cascade(ours)     &  \textbf{98.89} & \textbf{99.04} & \textbf{92.18} & \textbf{93.01}\\ 
    \bottomrule
    \end{tabular}
    \vspace{-1em}
\end{table}

\begin{table}[htbp]
    \centering
    \small
    \caption{Ablation study on the regression error of the fine-stage on the KITTI360Pose dataset.}\label{tab:regressor}
    \vspace{-1em}
    \begin{tabular}{l|cc}
    \toprule
    Method                  &  Train Error $\downarrow$ & Validation Error $\downarrow$ \\
    \midrule
    w/o cascade training              & 10.24 (+1.72) & 10.01 (+0.86)             \\
    w/o cross-attention      & 9.57 (+1.05)  & 9.56 (+0.41)              \\
    w/o confidence weighting & 9.02 (+0.50)  & 9.23 (+0.08)              \\
    \midrule
    Ours                    & \textbf{8.52} & \textbf{9.15}              \\
    \bottomrule
    \end{tabular}
    \vspace{-1em}
\end{table}

\noindent\textbf{\textit{Matcher.}}
% As the first step of the fine stage, the studied the training strategy of this matcher.
Following~\cite{sarlin2020superglue}, we take precision and recall as the the evaluation metric of the matcher.
With an identical matcher architecture, we investigate the impact of training strategy on the matcher performance.
% cascaded strategy and the joint strategy in~\cite{kolmet2022text2pos}.
The results are shown in Table~\ref{tab:exp_matcher}.
It can be seen that compared with joint training~\cite{kolmet2022text2pos}, our cascaded training achieves not only high precision and recall in the training set, but also stronger generalization on the validation set.
The results demonstrate that the cascade training strategy is able to mitigate the multi-task optimization difficulty.

\begin{figure*}[h]
    \centering
    \includegraphics[width=1.0\textwidth]{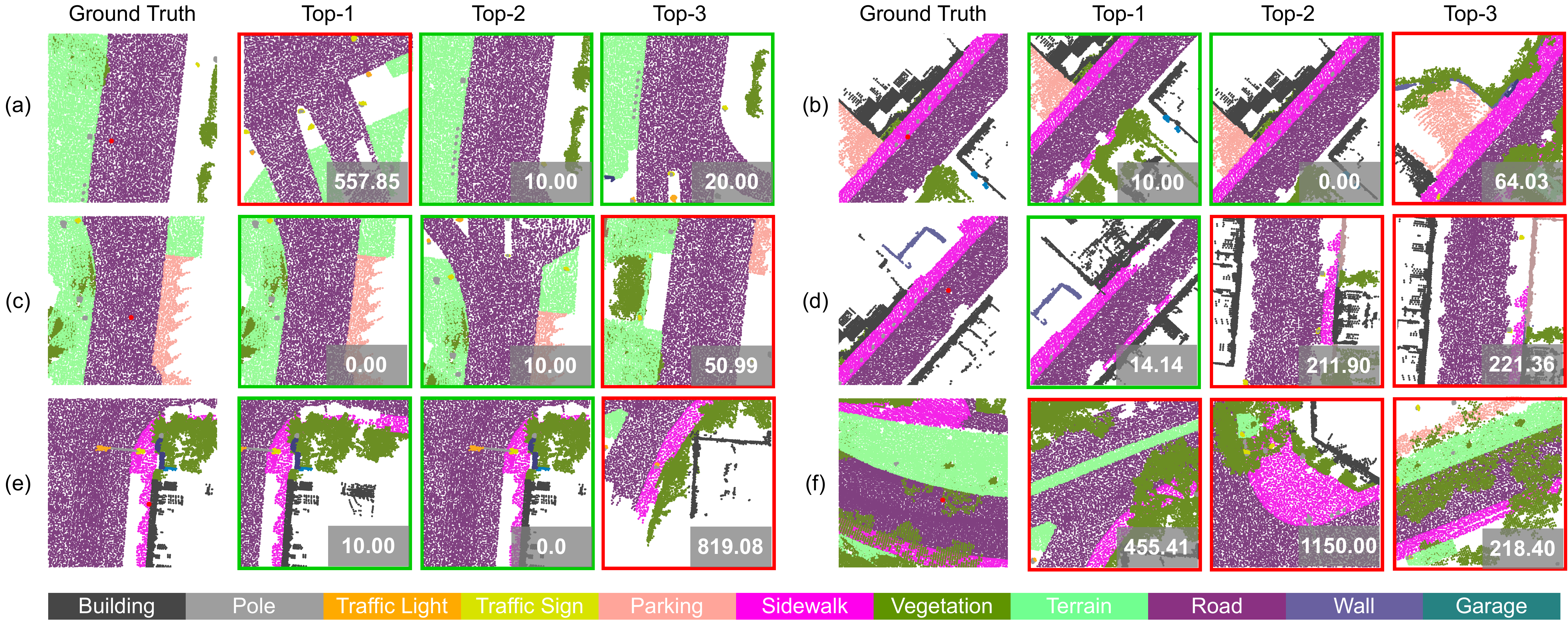}
    \caption{Qualitative retrieval results on KITTI360Pose validation set.
    The red dot in the ground truth cell indicates the target location.
    In each retrieved cell, the number in the lower right indicates the center distance between this cell and the ground truth. 
    Green box indicates positive cell which contains the target location, while red box indicates negative cells. 
    % where the center distance is less than 15$m$, while red boxes indicates negative cells. 
    }
    \label{fig:qualit_results}
\end{figure*}
% \vspace{-0.5em}
\noindent\textbf{\textit{Regressor.}}
The regressor predicts the target location based on the the matching results.
We study the effects of cascaded training, cross-attention based cross-modal fusion and confidence weighting for final location prediction.
We use regression error as evaluation metric and compare different versions on both KITTI360Pose training and validation set.
The results are shown in Table.~\ref{tab:regressor}.
Without cascaded training strategy, the regressor achieves an error of 10.24 and 10.01 on the training and validation set, respectively, which is 1.72 and 0.86 higher than that with cascaded training.
This result suggests that our cascaded training strategy also alleviates the optimization difficulty of the regressor, which was caused by the noisy intermediate results.
Furthermore, without cross-attention mechanism, the regression error also increases by a considerable margin, showing that cross-modal collaboration is important for precise location prediction.
Finally, with confidence-based weighting, we can further reduce the regression error on both the training and validation set, suggesting this information from the trained matcher can be further utilized to improve performance.

% \vspace{-2em}

% \vspace{-2em}
\subsection{Visualizations}
\vspace{-1em}

\begin{figure}[H]
    \centering
    \includegraphics[width=1.0\columnwidth]{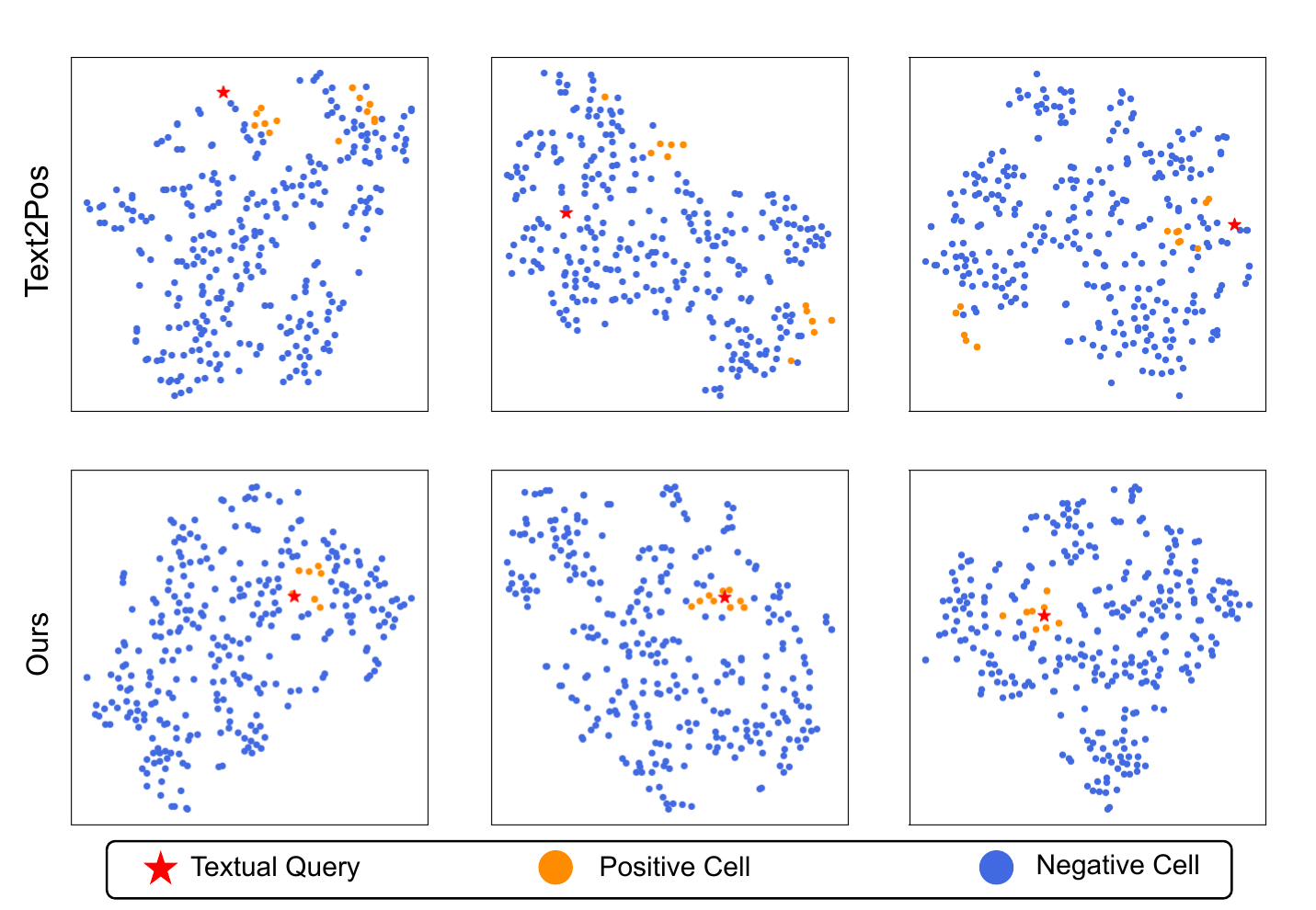}
    \vspace{-1em}
    \caption{T-SNE visualization of embedding space for the coarse stage.
    A cell is considered as positive if it contains the location described by the query.
    Compared with baseline method~\cite{kolmet2022text2pos}, our method can produce better representation where positive cells are closer to the target.}
    \label{fig:tsne}
\end{figure}

\noindent\textbf{Embedding Space Visualization.}
We visualize the learned embedding space via T-SNE~\cite{van2008visualizing} in Figure~\ref{fig:tsne}.
It can be observed that the baseline method Text2Pos~\cite{kolmet2022text2pos} results in a less discriminative space, where positive cells are relatively far away from the query and sometimes separated across the embedding space.
In contrast, our method draw positive cell and query representations closer in the embedding space, resulting in a more informative embedding space for retrieval.
% better retrieval performance.

\vspace{-0.5em}
\noindent\textbf{Qualitative Cell Retrieval Results.}
We show some example text to point cloud retrieval results in Figure.~\ref{fig:qualit_results}. 
For a given query, we visualize the top-3 retrieved cells. 
A retrieved cell is defined as positive if it contains the target location.
It can be observed that, our method can retrieve the ground truth cell or those close in most cases.
Sometimes, negative cells can also be retrieved,
\textit{e.g.}, top-1 in (a) and top-3 in (e).
It can be seen that these retrieved negative cells exhibit high semantic similarity with the ground truth cell, even though far away from it.
% and calls for the construction of a more discriminative representation. 
% and calls for richer representation for successful cell retrieval. 
We also show a failure case (f), where the retrieved cells are all negative.
It can be seen that even though far away from the target location, all these negative cells have instances similar to the ground truth. 
% but are far away from the target location.
These observations suggest that outdoor scenes are indeed of low diversity, indicating that successful retrieval requires highly discriminative representations to disambiguate the cells.  
% the cell representations should be further disambiguated to improve the performance. 
% This shows that the 
% richer information should be modeled for a more discriminative representation to disambiguate outdoor scenes.
% the top-3 retrieved results, even though sometimes far away from each other, often demonstrate high instance similarity. 
% This also shows outdoor scenes are often of low diversity, necessitating the incorporation of instance relations to increase the representation discriminability.

% even though the ground truth cell can successfully retrieved, the other negative cells,

\section{Conclusion}
In this work, we proposed a novel method for precise text-based localization from large-scale point clouds.
Our method employs a coarse-to-fine principle and pipelines this process into two stages.
For the coarse stage which is formulated as a textual query based cell retrieval task, 
we aim to improve representation discriminability for both point cloud and query representations.
This is achieved through explicit modeling of instance relations and implemented via a newly proposed Relation-Enhanced Transformer (RET).
The core of RET is a novel Relation-enhanced Self-Attention (RSA) mechanism, whereby the instance relations for the two modalities are explicitly incorporated into the value computation process in a unified manner.
% are modeled as spatial relation and feature fusions in visual and linguistic domain, respectively.
For the fine stage, our method performs description-instance matching and position refinement in a cascaded way, whereby cross-modal information collaboration is enhanced through the cross-attention mechanism.
Extensive experiments  on the KITTI360Pose dataset validated the effectiveness of the proposed method, which achieves new state-of-the-art performance.
Additional ablation studies further corroborate the effectiveness of each component in the proposed method.

\section{Acknowledgement}
This research is supported by the National Research Foundation,
Singapore under its Strategic Capability Research Centres Funding
Initiative. Any opinions, findings and conclusions or recommendations expressed in this material are those of the author(s) and do
not reflect the views of National Research Foundation, Singapore.

{
\bibliography{reference} 
}
\end{document}

\maketitle

\section{More about Data Pre-processing}
In this section, we introduce more details about the data pre-processing steps for the KITTI360Pose dataset~\cite{kolmet2022text2pos}.  
\subsection{Cell Division}
To properly handle the city-scale point clouds, we divide the point cloud into individual cells through the sliding window mechanism with a stride = 10$m$ of size 30$m$.
We also reject cells that does not have enough instances, so that our method can model better instance-wise relationships.
This cell division strategy is identical to that in~\cite{kolmet2022text2pos} for fair comparison.

\subsection{Stuff Classes}
In the point cloud, there can be \textit{stuff} classes. \textit{e.g.,} \texttt{terrain}, that can be wide or long in range and stretch across many adjacent cells.
To properly handle this, we perform clustering to divide such \textit{stuff} instances into multiple instances of the same class.
In this way, the \textit{stuff} class instances can also be viewed as \textit{thing} like instances, allowing unified instance processing in our framework.

\section{More implementation Details}
In this section, we elaborate more about the implementation details, which are omitted in the main text due to space limit.

\subsection{Instance Encoder}
The instance encoder embeds a point cloud instance $\Vec{P}_i \in \mathbb{R}^{N_i \times 6}$  into a vector $\Vec{p}_i \in \mathbb{R}^{d}$.
Here, $N_i$ means the number of points in the instance.  
We follow~\cite{kolmet2022text2pos} to also use the RGB color as part of feature, resulting in 6$d$ feature for each point.
The encoding process is achieved as follows:
\begin{equation}
\begin{split}
% \begin{align}
    & \Vec{p}_i = \textit{InstanceEncoder}(\Vec{P}_i) = \\
    & [
    \textit{PointNet++}(\Vec{P}_i); 
    \Mat{W}^{color}\Vec{a}_i; 
    \Mat{W}^{center}\Vec{c}_i;
    \Mat{W}^{count}\sigma(n_i)],
% \end{align}
\end{split}
\end{equation}
where $\Vec{a}_i \in \mathbb{R}^{3}$ is the average color in RGB format of the point cloud instance. 
This is important for the performance since color is often mentioned in the hints.
$\Vec{c}_i \in \mathbb{R}^3$ is the instance center coordinates. 
$n_i$ is a scalar that represents the number of points of the instance, which can be regarded as an approximation of the volume of the instance.
$\sigma(\cdot)$ is adopted to normalize the feature.
$\Mat{W}^*$ are learnable matrices that transform these features into the embedding space.
These features are concatenated together as the representation of instance $\Vec{p}_i$.

\subsection{Hint Encoder}
Since the hints in the KITTI360Pose dataset are generated using templates, we can directly encode the hints by parsing the hints.
We represent each hint into different groups: $\Vec{h}_j = \{w_0, w_1, ..., w_n\} = \{\Vec{w}_d, \Vec{w}_c, \Vec{w}_i\}$, where $\Vec{w}_d$ represents the word group for directions, $\Vec{W}_c$ for color group and $\Vec{w}_i$ for class group.
For example, given a hint \textit{The pose is east of a dark-green terrain.}, we directly parse it as $\Vec{w}_d = \{\textit{east} \}$, $\Vec{w}_c = \{\textit{dark-green}$\} and $\Vec{w}_i = \{ \textit{terrain}$\}.
Each word group is encoded by their learned word embeddings.
For cases there are multiple words in each group, we average the word embeddings to make the dimension consistent.
The obtained embeddings in each group is concatenated as the representation of this hint.

\subsection{Instance Padding for Hint-Instance Matching}
According to our data pre-processing procedure, there are 6 hints and indefinite number of instances in each cell.
To successfully adopt the SuperGlue~\cite{sarlin2020superglue} matcher, we need to fix the number of instances in each cell.
In our experiments, for cells with more than 16 instances, we randomly drop some cells.
While for cells with less than 16 instances, we randomly repeat some instances until the cell has 16 instances.
Note that this processing step is different from that in~\cite{kolmet2022text2pos}, which pad cells with random instances.
Our padding mechanism injects more meaningful information to the matching process, thereby better utilizing the information in each cell.

\subsection{Data Augmentation}
We also followed~\cite{kolmet2022text2pos} to adopt data augmentation strategies during training to improve the performance.
For the cell level augmentation, we randomly flip cells horizontally and/or vertically by flipping the direction word in the descriptions and the cell coordinates simultaneously.
% of the pose and instance center in each cell.
For example, given the direction word \texttt{east of}, we augment it by changing it to \texttt{west of} and flip the cell coordinates along the x-axis.
% The corresponding direction word, \textit{e.g.}, \textit{east of}, is also according changed.
Furthermore, we also perform fine-grained augmentation for both instances and hints.
For instance, we randomly rotate its points across z-axis and normalize the scale of all points to [0, 1] for both training and inference.
For descriptions, we randomly shuffle the hint order at each iteration. 

\subsection{PointNet++ Pretraining}
The pre-training procedure is identical to that in~\cite{kolmet2022text2pos} for pair comparison.
It is trained on the point cloud instance classification task on \textit{KITTI360}~\cite{Liao2021ARXIV_kitti360} dataset.
There are a total of 159,828 objects from 22 classes in the dataset. 
The model is optimized by an Adam~\cite{kingma2014adam} optimizer with a learning rate of $3e-3$ and a batch size of 32.

{
\bibliography{reference} 
}